\documentclass[conference]{IEEEtran}
\IEEEoverridecommandlockouts

\usepackage{cite}
\usepackage{amsmath,amssymb,amsfonts}
\usepackage{algorithm}
\usepackage[noend]{algpseudocode}
\usepackage{graphicx}
\graphicspath{{fig/}{./}} %
\usepackage{textcomp}
\usepackage{booktabs}
\usepackage{setspace}
\usepackage{multirow}
\usepackage[table]{xcolor}
\usepackage{subcaption}

\long\def\Skip#1{}  %

\providecommand{\PARstart}[2]{#1#2}

\title{AGRO-Nav: Autonomous Graph-based Orchard Navigation}

\author{\IEEEauthorblockN{Ho Young Yun, Jae Min Yu, and Duksu Kim}
\IEEEauthorblockA{School of Computer Engineering,\\
Korea University of Technology and Education (KOREATECH), Cheonan 31253, South Korea\\
Corresponding author: D.~Kim (\texttt{bluekds@koreatech.ac.kr})}
}

\begin{document}
\maketitle

\begin{abstract}
Orchards form semi-structured environments in which parallel tree rows create natural driving corridors, yet narrow inter-row clearance and dense foliage lead geometry-agnostic grid planners to drift off the row center and risk trunk or canopy contact. We present AGRO-Nav, an automated framework for static graph-based global planning in orchards. From tree-row lines fitted to trunk clusters in a SLAM point cloud, it builds, without any manual waypoints, a sparse topological graph of intra- and inter-row connectivity; a global route is then found by Dijkstra search on this graph, connected to the start and goal by any-angle Theta* segments, and smoothed with a cubic B-spline. In real-orchard trials, AGRO-Nav follows the row center with a mean error of about 0.08 m, far below the A* (0.31 m) and Theta* (0.43 m) shortest-path baselines, while planning roughly four to five times faster. In Isaac Sim, it attains the lowest error among A*, Theta*, and a reproduced RANSAC midline baseline and remains stable as tree density drops to 70\%, where the RANSAC baseline degrades. The resulting trajectories---straight row-centered segments joined by controlled turns---suit differential-drive and four-wheel-steering platforms.
\end{abstract}


\section{Introduction}
\label{sec:introduction}

\PARstart{O}{rchards} are a semi-structured navigation domain: parallel tree rows form natural corridors that should, in principle, make path planning easy. In practice, the corridors are narrow, the foliage is dense, and trees are frequently missing or unevenly spaced, so the row geometry that a robot must follow is only loosely defined. Geometry-agnostic planners cope poorly with these conditions. Generic occupancy-grid shortest-path planners such as A* and any-angle Theta* treat the orchard as undifferentiated free space; because they optimize for travel distance rather than row adherence, they cut corners at row transitions and can route a path straight through a tree row instead of staying centered in the corridor. Structural row-fitting methods that follow a RANSAC-estimated midline do respect the local row geometry, but they reason only over the trunks visible nearby and degrade as rows become sparse, when too few trunks remain to recover a continuous boundary. What is missing is an automatic pipeline that turns a SLAM map into a global, reusable representation of the orchard's row topology and plans over it directly, so that the resulting route is structure-aligned at the scale of an entire orchard block rather than locally fitted row by row.

Prior orchard navigation work has approached this from two directions, neither of which yields such a representation. The first follows rows locally without abstracting them: Underwood et al.~\cite{Underwood15} fuse dual 2D LiDAR scans into trunk centerlines, and Blok et al.~\cite{Blok19} achieve GPS-free row tracking by probabilistic filtering of LiDAR line features, but both stop at local row following rather than a global connectivity graph. The second uses graphs that are hand-built or implicit: Andersen et al.~\cite{Andersen10} manually construct an orchard graph for rule-based traversal, and Bertoglio et al.~\cite{Bertoglio23} navigate vineyards map-free by following only the two nearest rows, while end-to-end frameworks such as that of Wang et al.~\cite{Wang24} fuse LiDAR, IMU, and real-time kinematic GNSS (RTK-GNSS) for precise localization but do not expose a routable topology. The most closely related graph-based method is that of Jin et al.~\cite{Jin24}, who build a context-aware roadmap from a LiDAR-SLAM cloud using a learned semantic segmentation network, yet still specify the per-tree access nodes manually---by their own account because real orchard layouts are too irregular for fully automatic placement---and plan shortest paths rather than row-centered routes. For local row fitting, the RANSAC trunk-row midline method of Jiang and Ahamed~\cite{jiang2023navigation} is the most direct structural baseline. AGRO-Nav targets the gap both lines of work leave open.

This paper introduces AGRO-Nav, an autonomous graph-based orchard navigation framework, with four contributions. (C1) \textit{Automatic topology extraction}: from the PCA-fitted tree-row lines recovered from a single SLAM-generated 3D point cloud~\cite{Yun24}, AGRO-Nav automatically builds a sparse, connected topological graph of the orchard, requiring no manual waypoints. (C2) \textit{Static graph-based global planning}: it computes a globally row-aligned route by Dijkstra search over this graph, with any-angle Theta* entry and exit segments on a static 2D cost map. (C3) \textit{Smooth trajectory generation}: the route is smoothed with a cubic B-spline into controller-friendly trajectories suitable for both differential-drive and four-wheel-steering (4WS) platforms. (C4) \textit{Comprehensive evaluation}: we evaluate AGRO-Nav in real orchard field trials and in Isaac Sim---including a full-versus-70\% tree-density ablation---against A*, Theta*, and a reproduced RANSAC structural baseline, where it attains the lowest row-center error in every comparison and the fastest planning time. It also holds the largest minimum clearance from the trunks, with the least variation along the route---an orchard-safety property analyzed in Sec.~\ref{sec:results}.

\section{Related Work}\label{sec:related}

\subsection{Orchard Navigation and Mapping}

Row-structured perennial crops (apples, pears, citrus, grapes) form semi-structured environments---more repetitive and cluttered than urban roads, yet more regular than off-road terrain.
Early agricultural robots followed pre-mapped centerlines with RTK-GNSS waypoints~\cite{Noguchi01}, and later systems added LiDAR or vision for row detection and obstacle avoidance~\cite{Guerrero11,Bayar15,Underwood15}.
Recent work exploits trellis geometry to constrain the search space, pairing row-following controllers and headland-turn planners with behavior-tree mode switching~\cite{Colledanchise18}, and Blok et al.~\cite{Blok19} showed robust map-free row following from 2D LiDAR alone using particle and Kalman filters.

Perception in orchards is hampered by long repetitive corridors, sparse distant trunk returns, and canopy motion.
LiDAR(-inertial) odometry and SLAM (e.g., LOAM, LIO-SAM) give drift-resilient maps under vegetation and GNSS dropout~\cite{Zhang14,Shan20}, and camera--laser fusion has been used for orchard mapping and localization~\cite{Shalal15}; topology-oriented SLAM further extracts skeletons or graphs from point clouds~\cite{Bloechliger18}. Such an abstraction fits orchards particularly well: the free space is already a network of long, narrow corridors meeting only at the headlands, so reducing it to a graph discards little, whereas the same reduction in open space would throw away most of the drivable area.
On top of such maps, rows and centerlines are recovered by vision-based crop-row detection~\cite{Guerrero11}, LiDAR trunk clustering with RANSAC/PCA line fitting~\cite{Underwood15,jiang2023navigation}, or learned trunk detection from RGB-D imagery~\cite{Brown24}.
Among these, the RANSAC trunk-row midline method of Jiang and Ahamed~\cite{jiang2023navigation}---which groups the detected trunks into a left and a right row, fits a boundary line to each by RANSAC, and steers along the midline between them---is our most direct structural comparator and is reproduced as a baseline.

Despite this progress, most pipelines either assume manually curated routes or treat rows as purely local features, without abstracting them into a global, reusable topological representation; the interface from a raw SLAM map to a routable graph with cross-row and headland links is rarely made explicit.
In contrast, our method segments all tree rows automatically and builds a connected, parameterized row-graph, eliminating manual labeling and exposing tunable resolution and connectivity per block.

\subsection{Graph-Based Global Planning}

Topological and sampling-based planners (PRM, RRG/RRT*) trade optimality for scalability in large spaces~\cite{Kavraki96,Karaman11}, but in structured settings a problem-specific graph usually performs better: corridor roadmaps (Voronoi skeletons, medial axes) reduce branching and improve clearance~\cite{Choset00}.
For orchards, a row graph with headland connectors bounds search to semantically meaningful routes---within-row travel and controlled cross-row transitions.
Classical shortest-path solvers (Dijkstra/A*) then give deterministic global plans~\cite{Dijkstra59,Hart68}, any-angle variants (Theta*) shorten junction connectors~\cite{Nash07}, and inflation-based cost shaping preserves vegetation clearance~\cite{Nav225}.

Prior agricultural work often leaves graph construction manual or implicit (waypoints per row), limiting portability across diverse blocks.
The closest graph-based approach is Jin et al.~\cite{Jin24}, who derive a context-aware roadmap from LiDAR-SLAM point clouds: a learned semantic--instance segmentation network labels trees, ground, and obstacles, tree rows are recovered by line detection, and Dijkstra search returns the shortest heading-compliant path.
Two choices set our work apart.
First, their roadmap still requires manually specified per-tree access nodes---by the authors' own account, because real orchards are too irregular for fully automatic placement---and depends on a trained model, whereas our graph is built fully automatically from non-learned trunk geometry (PCA-fitted tree-row lines) with tunable node spacing and cross-row thresholds, needing neither labeled data nor manual nodes.
Second, their planner minimizes travel distance, while ours plans explicitly for row-center adherence.
These are complementary trade-offs: learned semantics help in visually complex scenes, whereas our geometric construction avoids training and manual labeling and keeps resolution and connectivity directly adjustable.
On the constructed graph our routes are optimal, though not in the continuous space.

\section{AGRO-Nav Framework}\label{sec:method}
In this section, we first provide a brief overview of the AGRO-Nav framework (Sec.~\ref{subsec:overview}) and then formalize the planning problem and notation (Sec.~\ref{subsec:formulation}).
We then explain the details of the two key components of our system: topological graph construction (Sec.~\ref{subsec:map_const}) and static graph-based global planning using the constructed graph (Sec.~\ref{subsec:path_planning}).

\begin{figure*}[!t]
    \centering
    \includegraphics[width=1.0\textwidth]{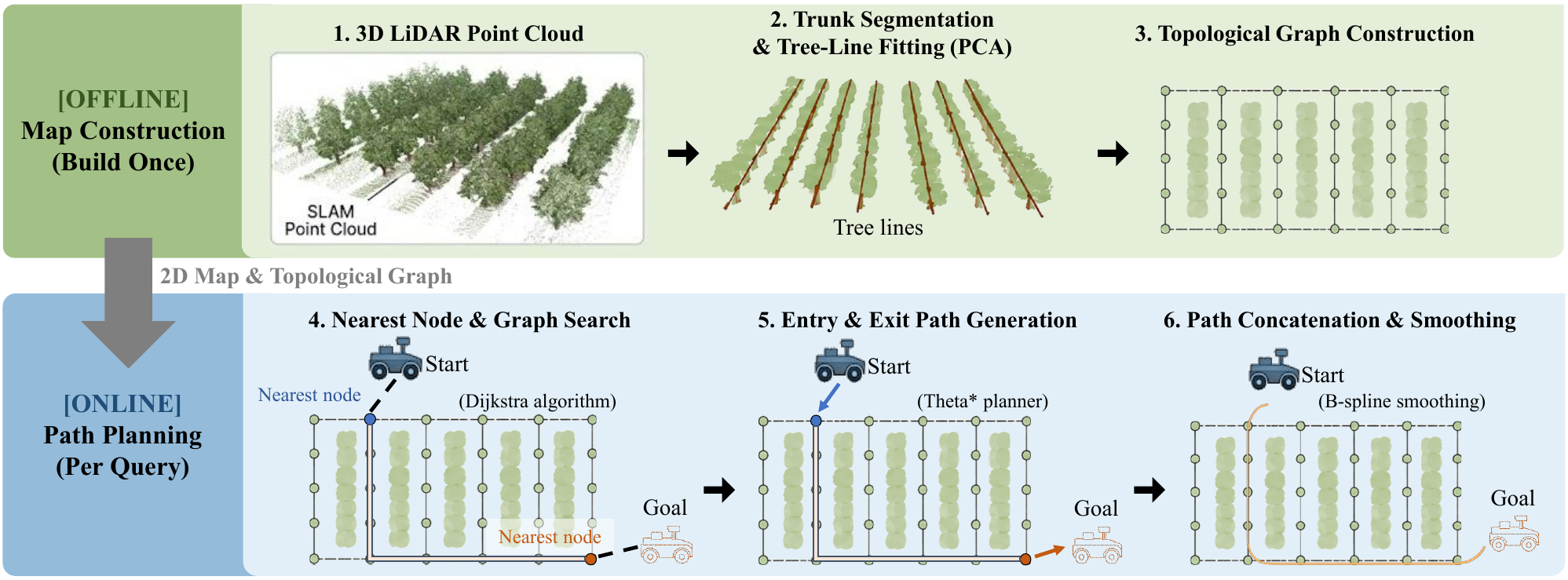}
\caption{Overview of the AGRO-Nav pipeline. An offline stage, built once per orchard map, converts a 3D SLAM point cloud into a 2D map with a topological graph; an online stage then plans a row-centered trajectory over that graph for each query. Sec.~\ref{subsec:overview} describes the stages.}
\label{fig1}
\end{figure*}

\subsection{System Overview}\label{subsec:overview}
The framework is divided into two primary stages: offline map construction (Fig.~\ref{fig1}, steps 1--3) and online path planning (steps 4--6).

\textbf{Map construction.}
The process begins with a 3D point cloud of the environment, typically generated by a LiDAR-based SLAM system.
Tree trunks are first segmented from this point cloud, and their positions are used to fit line models representing the orchard rows via PCA, following the method of~\cite{Yun24}.
Subsequently, graph nodes are sampled along the centerline between each pair of adjacent tree lines.
Edges are then established between nodes to capture the orchard's traversable structure, creating connections both along individual rows and across headlands to link adjacent rows (Sec.~\ref{subsec:map_const}).
The resulting node-edge network forms a compact and efficient representation, which we define as the \textit{topological graph}.

\textbf{Path planning.}
Once a navigation task is initiated with a start and goal pose, a single static global path is computed over the topological graph (Sec.~\ref{subsec:path_planning}).
The start and goal positions are first snapped to their nearest graph nodes, and a row-aligned route between these two nodes is obtained with Dijkstra's shortest-path search on the graph.
Because the actual start and goal poses generally lie off the graph, any-angle Theta* then generates short entry and exit segments that connect them to this route on a static 2D cost map, which enforces clearance from occupied cells.
The entry segment, graph route, and exit segment are concatenated and finally smoothed with a cubic B-spline and uniformly resampled into the output trajectory.
By routing over the graph in this way, the pipeline aligns the robot with the centerlines between tree rows, constraining the path to the orchard's traversable structure and yielding safe, efficient, row-centered traversal across entire orchard blocks.

\subsection{Problem Formulation and Notation}\label{subsec:formulation}
We formalize orchard global planning as the task of producing a single static, row-centered trajectory from a start pose to a goal pose, using a topological abstraction of the orchard structure as the intermediate representation.

\textbf{Inputs.}
The framework takes as input a SLAM-generated 3D point cloud of the orchard and the 2D occupancy/cost map derived from it.
From the perception front-end of~\cite{Yun24}, it receives a set of detected tree lines $L = \{\ell_i\}$, where each line $\ell_i$ is represented by its two endpoints $\mathbf{p}_i^{(1)}, \mathbf{p}_i^{(2)} \in \mathbb{R}^2$ on the ground plane.
A navigation request additionally provides a start pose and a goal pose in free space, together with an optional driving-region polygon $\Omega$ that bounds the permissible driving region.

\textbf{Output and objective.}
From these inputs the framework constructs a topological graph $G=(V,E)$ that abstracts the orchard's traversable structure---intra-row corridors and cross-row headland transitions---and returns a smooth, uniformly resampled, row-centered trajectory between the two poses.
It seeks to minimize lateral deviation from the inter-row centerlines (row adherence) and planning time, subject to static clearance from occupied cells and to remaining within $\Omega$.
Node placement is bounded by $\Omega$, cross-row edges require line-of-sight and a lateral separation below $\tau_{\text{row}}$, and every edge is capped at length $L_{\max}$; Sec.~\ref{subsec:map_const} formalizes these.
The trajectory must additionally be trackable by both differential-drive and 4WS platforms (Sec.~\ref{subsec:path_planning}).

\textit{Static-environment assumption:} the orchard is treated as a static, pre-mapped environment, and the planner produces a single global trajectory over this fixed structure. Avoidance of dynamic or moving obstacles is out of scope and is left to a downstream motion controller; the 2D cost map is used only to enforce static clearance and to search any-angle entry/exit segments, not to replan reactively at runtime.

\subsection{Topological Graph Construction}\label{subsec:map_const}

The graph is built from the tree-line set $L=\{\ell_i\}$ in two stages -- node generation and edge creation.

\textbf{Node generation.}
The detected tree lines are first projected onto the 2D ground plane.
Each line $\ell_i$ is defined by its endpoints, $\mathbf{p}_i^{(1)}, \mathbf{p}_i^{(2)} \in \mathbb{R}^2$, and its corresponding unit direction vector:
\begin{equation}
\mathbf{u}_i = \frac{\mathbf{p}_i^{(2)} - \mathbf{p}_i^{(1)}}{\left\lVert \mathbf{p}_i^{(2)} - \mathbf{p}_i^{(1)} \right\rVert}.
\end{equation}
Adjacent tree lines, $\ell_i$ and $\ell_j$, are then grouped to form a traversable row.
This grouping is the \textsc{GroupAdjacentLines} step in Algorithm~\ref{alg:map_construction}: two tree lines are paired into a row when (i) their orientations are sufficiently similar (the angle between $\mathbf{u}_i$ and $\mathbf{u}_j$ is below $25^{\circ}$) and (ii) they are laterally adjacent (their perpendicular separation lies within 3--5~m, on the order of the inter-row spacing).
For each row pair, we establish a central axis for node placement.
First, a seed point $\mathbf{s}_{ij}$ is computed as the geometric center of the four endpoints:
\begin{equation}  
\mathbf{s}_{ij} = \frac{\mathbf{p}_i^{(1)} + \mathbf{p}_i^{(2)} + \mathbf{p}_j^{(1)} + \mathbf{p}_j^{(2)}}{4}.
\end{equation}
Assuming the two tree lines are approximately parallel, we use the direction vector $\mathbf{u}_i$ to define the row's orientation.
Nodes $\mathbf{v}_k$ are then sampled bidirectionally from the seed point $\mathbf{s}_{ij}$ at an interval $d_{\text{node}}$ set to the intra-row tree spacing (e.g., 2.5~m):
\begin{equation}  
\mathbf{v}_{k} = \mathbf{s}_{ij} + k\,d_{\text{node}}\,\mathbf{u}_i, \qquad k \in \mathbb{Z}.
\end{equation}
A generated node $\mathbf{v}_k$ is added to the map only if it lies within the permissible driving region $\Omega$.
This area can be manually defined or, by default, is computed as the bounding box enclosing the row pair's endpoints, expanded by a safety margin (e.g., 2~m).

\textbf{Graph edge construction.}
First, longitudinal edges connect consecutive nodes within the same tree row.

Second, lateral (cross-row) edges connect nodes in adjacent rows.
An edge between a node $\mathbf{v}^{(i)}_{k}$ from row $i$ and a node $\mathbf{v}^{(j)}_{k'}$ from row $j$ is created only if it meets two criteria: (1) the straight-line path between the nodes does not intersect any detected tree lines (line-of-sight); and (2) the lateral distance between the nodes is within a threshold $\tau_{\text{row}}$ (e.g., 4~m).

The lateral distance is the component of $(\mathbf{v}^{(i)}_{k}-\mathbf{v}^{(j)}_{k'})$ perpendicular to row $j$, computed with the orthogonal projection matrix $\Pi_{\perp j}$:
\begin{equation}  
\Pi_{\perp j} = \mathbf{I} - \mathbf{u}_j\mathbf{u}_j^\top,
\end{equation}
where $\mathbf{u}_j$ is the unit direction vector of row $j$.
The condition for creating a lateral edge is then:
\begin{equation}  
\left\lVert \Pi_{\perp j}\big(\mathbf{v}^{(i)}_{k}-\mathbf{v}^{(j)}_{k'}\big)\right\rVert < \tau_{\text{row}}.
\end{equation}
Finally, each edge longer than a threshold $L_{\max}$ (e.g., 0.2~m) is recursively bisected at its midpoint until every edge is at most $L_{\max}$, giving the graph a uniform resolution for planning.
Even at this resolution the state space stays far smaller than a grid, because the graph discretizes only the routes---the row-centerlines and their headland connectors---whereas an occupancy grid discretizes the entire orchard area.

Algorithm~\ref{alg:map_construction} summarizes the construction.

\begin{algorithm}[t]
\footnotesize
\caption{Topological Graph Construction}\label{alg:map_construction}
\begin{algorithmic}[1]
\Require
    Set of tree lines $L$; 
    Parameters $\Theta = \{d_{\text{node}}, \tau_{\text{row}}, L_{\max}\}$
\Ensure Graph $G=(V, E)$
\Function{ConstructTopologicalMap}{$L, \Theta$}
    \State $V, E \gets \emptyset, \emptyset$
    \State $R \gets \text{GroupAdjacentLines}(L)$
    \Statex \textit{\# Phase 1: Node Generation}
    \For{each row $r = (\ell_i, \ell_j) \in R$}
        \State $\mathbf{s} \gets (\mathbf{p}_i^{(1)} + \mathbf{p}_i^{(2)} + \mathbf{p}_j^{(1)} + \mathbf{p}_j^{(2)}) / 4$
        \State $\mathbf{u} \gets \text{GetDirection}(\ell_i)$
        \State $\Omega \gets \text{DefineBoundingBox}(r)$
        \For{$k \in \mathbb{Z}$ until nodes leave $\Omega$}
            \State $\mathbf{v} \gets \mathbf{s} + k \cdot d_{\text{node}} \cdot \mathbf{u}$
            \If{$\mathbf{v} \in \Omega$}
                \State $V \gets V \cup \{\mathbf{v}\}$
                \State \text{Associate a row ID with } $\mathbf{v}$
            \EndIf
        \EndFor
    \EndFor
    \Statex \textit{\# Phase 2: Edge Generation}
    \For{each row $r \in R$}
        \State $V_r \gets \text{GetNodesForRow}(V, r)$, sorted along the row direction
        \For{$k = 1$ to $|V_r|-1$}
            \State $E \gets E \cup \{ (V_r[k], V_r[k+1]) \}$
        \EndFor
    \EndFor

    \For{each pair of adjacent rows $(r_i, r_j)$}
        \State $\mathbf{u}_j \gets \text{GetDirection}(r_j)$;\quad $\Pi_{\perp j} \gets \mathbf{I} - \mathbf{u}_j \mathbf{u}_j^\top$
        \For{each node $\mathbf{v}_i$ in row $r_i$}
            \For{each node $\mathbf{v}_j$ in row $r_j$}
                \If{$\|\Pi_{\perp j}(\mathbf{v}_i - \mathbf{v}_j)\| < \tau_{\text{row}}$ \textbf{and} $\text{NoIntersection}(\{\mathbf{v}_i, \mathbf{v}_j\}, L)$}
                    \State $E \gets E \cup \{ (\mathbf{v}_i, \mathbf{v}_j) \}$
                \EndIf
            \EndFor
        \EndFor
    \EndFor

    \Statex \textit{\# Phase 3: Edge Subdivision}
    \State $(V, E) \gets \text{BisectEdges}(V, E, L_{\max})$ \Comment{split at midpoints until every edge $\le L_{\max}$}

    \State \Return $G=(V, E)$
\EndFunction
\end{algorithmic}
\end{algorithm}

\subsection{Static Graph-based Global Planning}\label{subsec:path_planning}

Given a start and a goal pose, a single static trajectory is computed over the precomputed topological graph.
A 2D cost map, built once from the occupancy grid, serves only the static-clearance and entry/exit roles stated in Sec.~\ref{subsec:formulation}.
Unlike conventional approaches that use exponential-decay cost functions~\cite{Nav225}, we apply a linear distance-based cost function within a maximum influence distance $d_{\max}$ (e.g., 1~m): the cost is maximal at occupied cells and decreases linearly to zero at $d_{\max}$.
This produces a more uniform cost gradient around obstacles, encouraging the planner to keep a safe clearance.

\textbf{Graph-level path planning.}
The start and goal poses are snapped to their nearest nodes $\mathbf{v}_{\text{start}}$ and $\mathbf{v}_{\text{goal}}$ on $G=(V,E)$, and the route between them is computed with Dijkstra's algorithm~\cite{Dijkstra59}, each edge weighted by its Euclidean length.

\textbf{Entry and exit path planning.}
Because the start and goal generally lie off the graph, the Theta* planner~\cite{Nash07} computes the entry and exit segments on the 2D cost map.
We select Theta* over standard A* and Hybrid A* because its ``any-angle,'' line-of-sight search produces smoother paths than grid-aligned planners.

\textbf{Path smoothing and resampling.}
Concatenating the entry, graph-level, and exit paths leaves sharp turns at the junction nodes $\mathbf{v}_{\text{start}}$ and $\mathbf{v}_{\text{goal}}$; we therefore apply cubic B-spline smoothing to the entire waypoint sequence.
The result is a smooth trajectory of straight in-row segments joined by controlled turns at row transitions, trackable by both differential-drive and 4WS platforms.
Finally, the trajectory is uniformly resampled at a fixed spatial interval (e.g., 0.05~m) for a consistent waypoint density.

\section{Results and Analysis}\label{sec:results}

\subsection{Experimental Setup}\label{subsec:setup}

Across both settings, we compare AGRO-Nav against Nav2's default A*~\cite{Hart68} and Theta*~\cite{Nash07}, with an additional RANSAC-based structural baseline~\cite{jiang2023navigation} included in the simulation experiments, where an exact 3D reference is available. Each evaluated route is given as a start pose, a goal pose, and intermediate waypoints selecting the rows to traverse; without them a shortest-path planner would simply cut straight to the goal. The same waypoints are supplied to every planner, and they constrain only which route is requested, not how the graph is built. To ensure a fair comparison, all planners are executed on the same 2D occupancy grid under identical conditions, and each planner is run ten times per start--goal pair; the path-quality metrics are averaged over the evaluated start--goal pairs, while the planning time is averaged over the ten repetitions to reduce runtime fluctuation. We evaluate each planner's output path---its geometry, row-center adherence, and planning time---computed offline on the recorded and simulated occupancy maps. For AGRO-Nav, the reported planning time covers only the online query (nearest-node lookup, Dijkstra graph search, Theta* entry/exit search, and path smoothing; Sec.~\ref{subsec:path_planning}); the topological graph is built once per orchard map during the offline mapping stage, and this one-time cost is amortized over all subsequent queries.

In every experiment, the accuracy metric is the mean lateral deviation from the orchard row center: for each sampled reference row-center point, the error is the minimum Euclidean distance to the generated path. Clearance is sampled at the same points, taken from where the path crosses each one to the nearest trunk, so all planners are compared at identical positions along the rows. Because ground-truth trajectories are unavailable, the reference row-centerlines serve as a pseudo-ground-truth, so the metric measures adherence to the row center rather than path optimality. As AGRO-Nav is explicitly designed to follow row centers while the A* and Theta* baselines optimize for the shortest path, this reference structurally favors a row-centering planner; the reported errors should therefore be interpreted with this bias in mind rather than as a measure of general path optimality.

\begin{figure}[t]
  \centering
  \begin{subfigure}[b]{0.485\columnwidth}
    \centering
    \includegraphics[width=\linewidth]{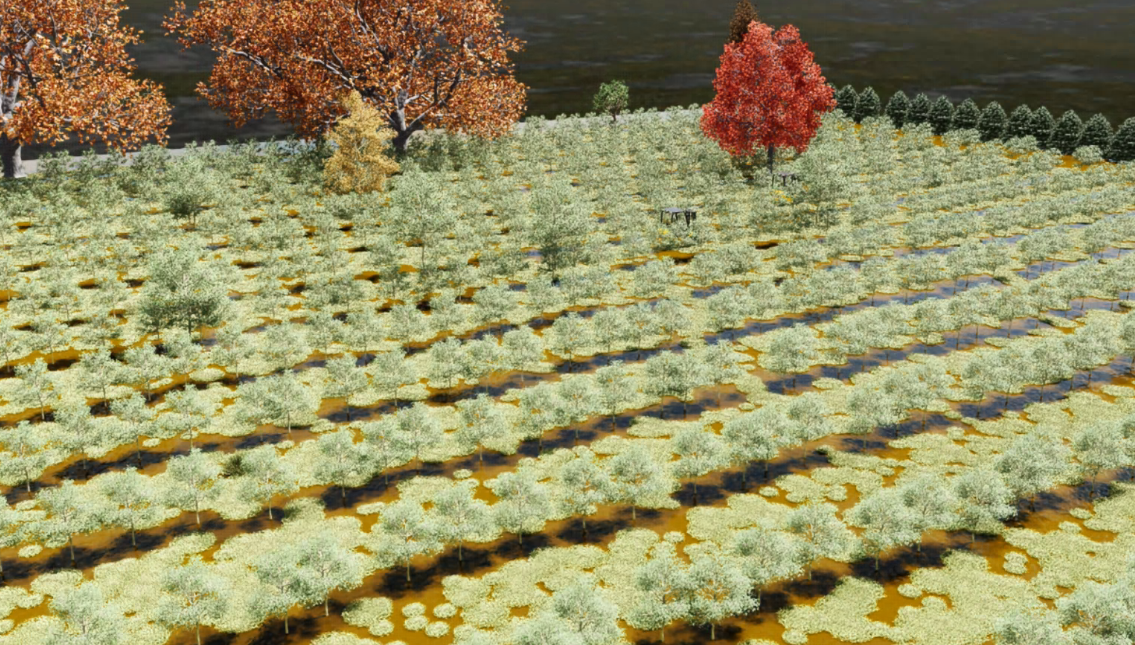}
    \caption{Overall environment.}
    \label{fig:sim_env_a}
  \end{subfigure}
  \hfill
  \begin{subfigure}[b]{0.485\columnwidth}
    \centering
    \includegraphics[width=\linewidth]{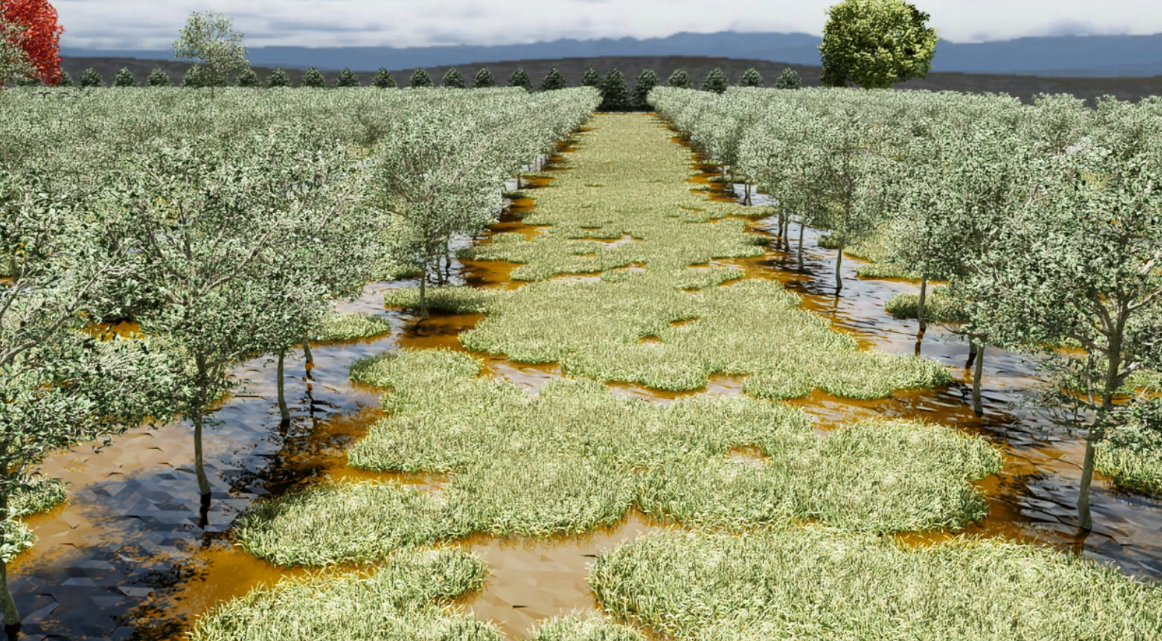}
    \caption{Structured orchard rows.}
    \label{fig:sim_env_b}
  \end{subfigure}
  \caption{Isaac Sim orchard environment used for the simulation-based validation.}
  \label{fig:sim_env}
\end{figure}

\textbf{Simulation setup.}
We first validate the pipeline in Isaac Sim under controlled, repeatable conditions. We model a robot matching the real 4WS field platform and construct an orchard environment whose structured tree rows reproduce the corridor-like geometry of real orchards (Fig.~\ref{fig:sim_env}), with average inter-row and intra-row spacings of approximately 4.0~m and 2.5~m. We evaluate six start--goal pairs, and six representative rows are each sampled at 11 points, giving 66 evaluation points; because the 3D tree positions are known, the reference row-centerlines are computed as the mid-positions between adjacent tree rows in 3D and projected onto the 2D planning map. To probe robustness to sparsity, a 70\% tree-density condition is generated by randomly removing 30\% of the trees while preserving the overall row layout.

\textbf{Real-world setup.}
We then evaluate on ROS2 datasets collected from a commercial orchard in Daegu, South Korea. The platform is a 4WS autonomous orchard-inspection robot equipped with a 128-channel LiDAR, RTK-GNSS, and an IMU, with all computation performed on an NVIDIA Jetson AGX Orin. We evaluate three representative start--goal pairs; since no ground truth exists, we manually extract the orchard's row-centerlines as the reference, dividing each of the three rows into ten segments (11 points per row) for 33 evaluation points in total.

\subsection{Simulation-based Validation}\label{subsec:sim_validation}

\begin{table*}[!t]
  \centering
  \caption{Quantitative comparison of path planning methods across the three evaluation settings.
  Error is reported as mean ($\pm$ standard deviation)\,/\,max lateral deviation from the row center.
  Clearance is reported as mean ($\pm$ standard deviation)\,/\,minimum distance from the planned path to the nearest tree trunk, where trunk positions are known (larger is safer).
  The RANSAC baseline requires exact 3D tree positions and is evaluated only in simulation.
  In each setting, bold marks the best value of each reported metric.}
  \label{tab:main_result}
  \setlength{\tabcolsep}{6pt}
  \small
  \begin{tabular}{lcccc}
    \hline
    \textbf{Method} & \textbf{Mean ($\pm$ Std) / Max Error [m]} & \textbf{Length [m]} & \textbf{Mean ($\pm$ Std) / Min Clearance [m]} & \textbf{Time [ms]} \\ \hline
    \hline
    \multicolumn{5}{c}{\textit{Simulation environment (full tree density)}} \\ \hline
    A*       & 0.95 ($\pm$ 0.43) / 1.94          & 509.66          & 1.96 ($\pm$ 0.38) / 1.26          & 61.95 \\
    Theta*   & 0.79 ($\pm$ 0.54) / 2.22          & \textbf{499.33} & 2.07 ($\pm$ 0.42) / 1.01          & 69.16 \\
    RANSAC   & 0.18 ($\pm$ 0.10) / 0.47          & 524.30          & \textbf{2.58 ($\pm$ 0.27)} / 1.86 & 33.74 \\
    AGRO-Nav & \textbf{0.14 ($\pm$ 0.10) / 0.42} & 527.29          & 2.57 ($\pm$ 0.25) / \textbf{1.93} & \textbf{14.24} \\
    \hline
    \multicolumn{5}{c}{\textit{Simulation environment (70\% tree density)}} \\ \hline
    A*       & 0.90 ($\pm$ 0.55) / 2.31          & 515.56          & 1.98 ($\pm$ 0.34) / 1.24          & 62.30 \\
    Theta*   & 0.94 ($\pm$ 0.66) / 2.45          & \textbf{506.17} & 1.96 ($\pm$ 0.49) / 1.02          & 68.94 \\
    RANSAC   & 0.24 ($\pm$ 0.16) / 0.54          & 530.74          & 2.49 ($\pm$ 0.27) / 1.86          & 34.54 \\
    AGRO-Nav & \textbf{0.14 ($\pm$ 0.10) / 0.44} & 533.19          & \textbf{2.57 ($\pm$ 0.24) / 2.03} & \textbf{13.48} \\
    \hline
    \multicolumn{5}{c}{\textit{Real orchard}} \\ \hline
    A*       & 0.31 ($\pm$ 0.15) / 0.73          & 214.09          & 1.63 ($\pm$ 0.35) / 0.97          & 51.24 \\
    Theta*   & 0.43 ($\pm$ 0.20) / 0.81          & \textbf{212.70} & \textbf{1.89 ($\pm$ 0.36)} / 1.19 & 58.48 \\
    AGRO-Nav & \textbf{0.08 ($\pm$ 0.07) / 0.22} & 218.03          & 1.85 ($\pm$ 0.27) / \textbf{1.29} & \textbf{11.52} \\
    \hline
  \end{tabular}
\end{table*}

\begin{figure*}[!t]
  \centering
  \begin{subfigure}[b]{\textwidth}
    \centering
    \includegraphics[width=\linewidth]{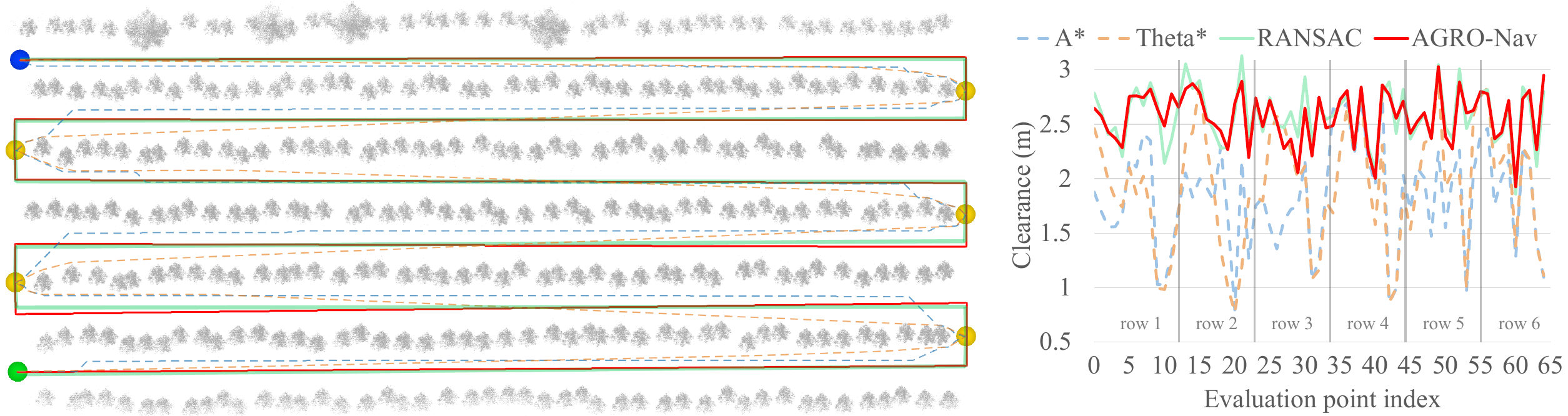}
    \caption{Full tree density (100\%).}
    \label{fig:sim_full_density_result}
  \end{subfigure}
  \\[4pt]
  \begin{subfigure}[b]{\textwidth}
    \centering
    \includegraphics[width=\linewidth]{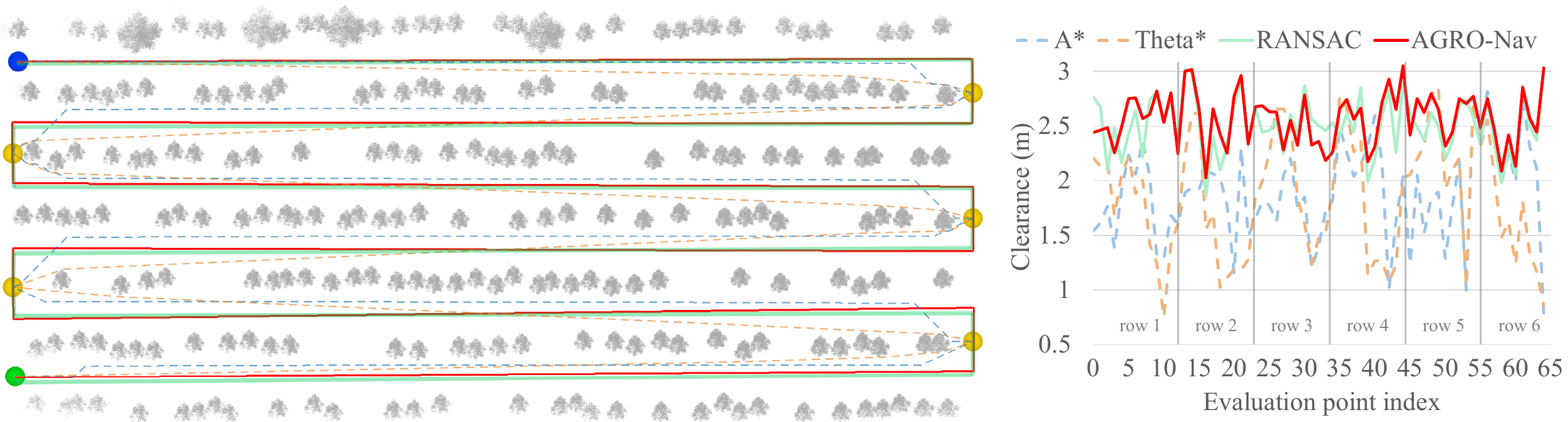}
    \caption{70\% tree density.}
    \label{fig:sim_density_70_result}
  \end{subfigure}
  \caption{Simulation results at full and 70\% tree density. Each panel pairs the planned paths on the 2D Isaac Sim orchard map (left) with the trunk clearance along those paths (right; higher is safer). On the maps, green and blue mark the start and goal of each evaluated route and yellow the intermediate waypoints that select the rows to traverse.}
  \label{fig:sim_density_maps}
\end{figure*}

The goal of the simulation study is to test whether AGRO-Nav can automatically extract the orchard's row topology and plan row-centered paths through graph-based search, without any hand-placed graph nodes---a capability that simulation is well suited to isolate, since the known tree geometry yields an exact row-center reference rather than the manually traced one available in the field. We compare A*, Theta*, a RANSAC-based structural baseline, and AGRO-Nav under two tree-density conditions: the full-density environment and a 70\% tree-density environment. The RANSAC baseline follows the trunk-row midline strategy of~\cite{jiang2023navigation} described in Sec.~\ref{sec:related}. Since the original source code was not publicly available, we reproduced it from the published algorithmic description as closely as possible; the result should therefore be read as a representative structural row-fitting baseline rather than an official implementation of the original method.

Table~\ref{tab:main_result} collects the quantitative results for all three evaluation settings; the two simulation blocks (full and 70\% tree density) are discussed here and the real-orchard block in Sec.~\ref{subsec:real_world_eval}. In the two simulation blocks, the reported mean and maximum errors were computed over the 66 sampled evaluation points defined in Sec.~\ref{subsec:setup}. The point-wise clearance along the resulting paths is visualized in Fig.~\ref{fig:sim_density_maps}. Beyond averaged accuracy, each error entry also reports the standard deviation over those same points, capturing how consistently a planner stays centered along the whole route rather than only on average. AGRO-Nav shows the smallest dispersion among the compared methods (0.10~m at both densities, against 0.43--0.66~m for the grid-based planners), indicating that its low mean error reflects uniformly centered adherence rather than an average over widely varying deviations. In the full-density condition, AGRO-Nav achieved the lowest mean error of 0.14 m and the lowest maximum error of 0.42 m among all compared methods. This result indicates that the extracted topological graph effectively preserves the row-center structure of the orchard. Compared with A* and Theta*, AGRO-Nav generated paths that better followed the intended row-centered navigation pattern. Although the RANSAC-based method also produced a low mean error, AGRO-Nav showed the best row-center adherence while maintaining the graph-based path structure required for autonomous orchard navigation. The clearance entries report how close each path comes to the nearest trunk, both on average over the route and at its single worst point. Safety is governed by that worst point, and AGRO-Nav holds the largest minimum clearance in every setting while varying the least along the route (0.24--0.27~m standard deviation against 0.34--0.49~m for the grid-based planners), so its margin is uniform rather than occasionally generous. Theta* comes within 1.01~m of a trunk at full density, and under the 70\% condition its path visibly cuts through a tree row (Fig.~\ref{fig:sim_density_70_result}). The RANSAC baseline, which also targets the row center, reaches a comparable mean clearance but a lower minimum at both full density (1.86~m versus 1.93~m) and 70\% density (1.86~m versus 2.03~m), the gap widening as the rows thin out, consistent with its rising mean error there. AGRO-Nav is also the fastest planner in both conditions (14.24~ms and 13.48~ms), roughly four to five times quicker than the grid-based planners and 2.4--2.6$\times$ quicker than RANSAC, so its row-centered clearance is obtained at a fraction of the query cost of the one baseline that matches it. We return to this safety advantage in Sec.~\ref{subsec:real_world_eval}.

The 70\% density condition was additionally tested to examine whether the proposed method remains stable when the row structure becomes partially sparse. In this setting, AGRO-Nav maintained a mean error of 0.14 m, essentially unchanged from the full-density result, with only marginal increases in maximum error and path length, and its row-centerline error remained substantially lower than those of the grid-based planners.

This stability follows from the graph construction process: rather than deriving the path from isolated tree observations, AGRO-Nav first captures the global row structure and then builds the graph from the extracted row geometry. As long as the remaining trunks provide enough evidence for that structure, the resulting graph stays close to the full-density one.

Compared with the full-density condition, the grid-based planners showed different error tendencies in the 70\% density environment. A* achieved a lower mean error than in the full-density case, but its average deviation remained much larger than that of AGRO-Nav. Theta* also showed a large mean error and a high maximum error, indicating that its shortest-path-oriented behavior can still lead to substantial deviations from the row-center reference. In particular, in the third row from the bottom of the 70\% density map, Theta* generated a path that penetrated through a tree row rather than following the intended row corridor. Such behavior is undesirable for orchard navigation because the robot should remain within the safe traversable corridor between tree rows.
The RANSAC-based method showed a relatively small error, but its mean error increased from 0.18 m to 0.24 m when the tree density was reduced. This suggests that the RANSAC-based structural baseline is more sensitive to sparse tree distributions than the proposed topology-based approach. In additional exploratory trials in which the tree density was reduced below 70\% (conditions not included in the quantitative comparison of Table~\ref{tab:main_result}), line recognition became unstable in some regions because the remaining trees were too sparse to provide a continuous row structure. This indicates that AGRO-Nav, like other row-structure-based methods, requires a minimum level of row observability and may not operate reliably in extremely sparse orchard sections.

Fig.~\ref{fig:sim_density_maps} localizes what the table can only summarize: A* and Theta* drop sharply at row transitions and wherever the shortest path leaves the row structure, repeatedly falling near 1~m, whereas AGRO-Nav holds a nearly flat band around 2.5~m along the whole route and RANSAC tracks that band but dips further at the transitions. Both patterns persist at 70\% density, alongside the Theta* row penetration and the RANSAC error rise discussed above, confirming that AGRO-Nav does not depend on a perfectly dense tree-row structure.

\subsection{Real-world Orchard Evaluation}\label{subsec:real_world_eval}

\begin{figure}[!t]
  \centering
  \includegraphics[width=\columnwidth,keepaspectratio]{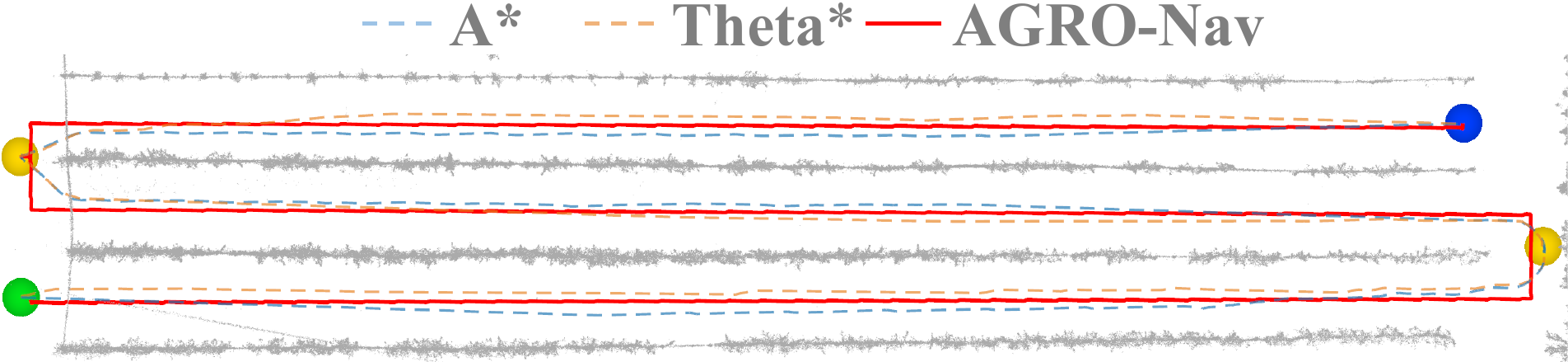}
  \caption{Paths generated by the three planners in the real orchard. Green and blue mark the start and goal, and yellow the intermediate waypoints.}
  \label{fig:real_world_result}
\end{figure}

The real-orchard block of Table~\ref{tab:main_result} reports the field metrics. AGRO-Nav achieves a mean error of only 0.08 m, confirming its ability to precisely follow the row-centerlines, and is approximately four to five times faster than the baselines (11.52 ms versus 51.24 ms and 58.48 ms). This speedup follows directly from planning over the small precomputed topological graph, whose nodes and edges are far fewer than the occupancy-grid cells that A* and Theta* must search, so the global route is found with substantially less effort.

While the total path length does not differ appreciably among the planners, AGRO-Nav results in a slightly longer trajectory (218.03 m versus 214.09 m for A* and 212.70 m for Theta*). This is an intended consequence of the design: rather than tracing the globally shortest path, AGRO-Nav follows straight row-centered segments joined by controlled turns at row transitions, which keeps the robot centered between rows and clear of the trunks. In contrast, baseline planners such as A* and Theta* do not account for the orchard's structural constraints and instead seek the shortest path over the 2D grid, leading to less structured, geometry-agnostic trajectories that pass closer to the trees, as illustrated in Fig.~\ref{fig:real_world_result}. We have frequently observed such deviations while operating the platform in the orchard, which motivated the proposed topology-based planning strategy.

This row-following motion pattern is also safer for orchard operation. Because AGRO-Nav turns only at the headlands, its trajectories keep a consistent distance from the trunks and surrounding canopy, and that clearance acts as a robustness margin: because the planned path stays well away from the trunks, the ordinary tracking error that any platform exhibits at operating speed is absorbed by this margin rather than translating into trunk contact. Both differential-drive and 4WS platforms turn in place or with a small radius, so they execute the controlled headland turns directly and preserve this safety margin in practice; grid-based planners, by contrast, tend to produce wide, continuously varying headland turns that are harder to execute cleanly and, in our field experience, can disturb the soil around the trees.

\section{Conclusion}
In this paper, we have presented AGRO-Nav, an automated graph-based global planning pipeline tailored to the structured yet challenging environment of orchards. Building on PCA-fitted tree-row lines recovered from a SLAM-generated 3D point cloud, AGRO-Nav constructs a connected topological graph automatically, thereby eliminating manual waypoint design. A global route is then planned by coupling Dijkstra routing over this sparse graph with Theta* any-angle entry and exit search on a static 2D cost map, which is markedly more efficient than searching the full occupancy grid. Finally, cubic B-spline smoothing and uniform resampling produce continuous trajectories of straight in-row segments joined by controlled turns at row transitions, suited to both differential-drive and 4WS platforms.

Field trials in a real orchard show that AGRO-Nav follows the row-center reference far more closely than the baselines (0.08 m mean row-center error versus 0.31 m for A* and 0.43 m for Theta*) while planning roughly four to five times faster (11.52 ms versus 51.24 ms and 58.48 ms). Complementary Isaac Sim experiments confirm the lowest mean and maximum row-center error among all compared methods at both full and 70\% tree density (0.14 m mean error at both densities). Together, these results connect high-level structural reasoning with smooth trajectory generation, enabling repeatable, efficient, and structure-aligned global planning in large-scale orchards.

\textbf{Limitations and Future Work.}
AGRO-Nav also has limitations that point to future work. Currently, only static global planning is evaluated: the orchard is treated as a static, pre-mapped environment, and avoidance of dynamic or moving obstacles is left to future work. In addition, the real-world evaluation was conducted at a single commercial orchard site. Future work will address these aspects by integrating a runtime local obstacle layer for dynamic environments, validating trajectory tracking and kinematic feasibility on the physical platform, and extending the evaluation to broader orchard layouts, seasons, and multiple sites.

\bibliographystyle{IEEEtran}
\bibliography{ref}

\end{document}